\documentclass{article}

\usepackage{microtype}
\usepackage{graphicx}
\usepackage{subfigure}
\usepackage{booktabs} % for professional tables

\usepackage{hyperref}

\usepackage[accepted]{icml2019}

\icmltitlerunning{Improved Tree Search for Automatic Program Synthesis}

\begin{document}

\twocolumn[
\icmltitle{Improved Tree Search for Automatic Program Synthesis}



\begin{icmlauthorlist}
\icmlauthor{Aran Carmon}{tau}
\icmlauthor{Lior Wolf}{tau}
\end{icmlauthorlist}

\icmlaffiliation{tau}{The School of Computer Science , Tel Aviv University}

\icmlcorrespondingauthor{Lior Wolf}{liorwolf@gmail.com}
\icmlcorrespondingauthor{Aran Carmon}{arancarmon@mail.tau.ac.il}

\icmlkeywords{Automatic Program Synthesis, Exploration, MCTS}

\vskip 0.3in
]



\printAffiliationsAndNotice  % leave blank if no need to mention equal contribution

\begin{abstract}
In the task of automatic program synthesis, one obtains pairs of matching inputs and outputs and generates a computer program, in a particular domain-specific language (DSL), which given each sample input returns the matching output. A key element is being able to perform an efficient search in the space of valid programs. Here, we suggest a variant of MCTS that leads to state of the art results on two vastly different DSLs. The exploration method we propose includes multiple contributions: a modified visit count, a preprocessing procedure for the training dataset, and encoding the part of the program that was already executed. 
\end{abstract}

\section{Introduction}
\label{sec:intro}

Search is a key part of many machine learning tasks, in which the output is a  sequence. We study the specific task of automatic program synthesis, given a specification in the form of input/output pairs~\cite{deepcoder,neuralguided, Gulwani:2016:PEA:2960761.2960764, predicting-a-correct-program-in-programming-by-example}. In this sequence generation task, similar to other generation tasks, such as machine translation and image captioning, there are multiple correct answers.

Different from most sentence generation tasks in NLP, in this task, one is able to directly evaluate the correctness of the output, by running the generated program. This leads to a well-defined and natural reward, when viewed as a reinforcement learning problem~\cite{bunel2018leveraging}: either the generated program produces the specified outputs given the matching inputs or not. Other variants of this reward may consider, for example, the length of the program, encouraging the generated program to be more efficient.

However, as reported by previous work~\cite{zohar2018automatic,chen2018execution}, employing a reinforcement learning approach, as opposed to training using a maximum likelihood loss to generate the single program that is available as the ground truth, either hurts performance or leads to a small increase in performance. This is despite training the MLE approach in a teacher-forcing way, in which, during training and unlike during test time, the partial programs considered are the prefix of the ground truth programs.

In this work we, therefore, focus on the MLE approach and consider different tree search strategies. These methods include the classical beam search, which is often used in the program synthesis domain, the CAB variant of it, which was used successfully in the past~\cite{zohar2018automatic}, and various MCTS approaches that we develop and explore.

In addition to studying the specific building blocks of the application of MCTS to this problem, we also suggest two other improvements. The first is a pre-processing step that is applied to the training set, and the second is the addition of an encoder of the current program's history. We demonstrate how these techniques can improve the obtained accuracy, both separately and when combined.

\section{Related Work}

We experiment with two DSLs. The first is the DeepCoder DSL~\cite{deepcoder,zohar2018automatic}, in which integer registers are being manipulated. Each register may contain either an integer or a list of integers, and the program may apply on the registers,  functions such as sort, tail, and multiply.~\citet{deepcoder} have used the predictions of a neural network that was applied to the input/output pairs, in order to augment classical program search techniques, such as SMT-solvers and enumerative search. Specifically, the neural network produced a vector of probabilities for the existence of every command or function in the program.  Later on,~\citet{zohar2018automatic} used a neural network to predict the probability of the next statement given the execution state, after applying the partial program as part of a beam search. Specifically, the CAB search strategy~\cite{cab} was used. Since, in this DSL, every statement returns one integer or list into the memory, and since this memory is limited, a learned garbage collection mechanism was used to discard the results of previous statements from the memory. Our work on this DSL extends the method of~\cite{zohar2018automatic}, and also employs this garbage collector.

The Karel DSL~\cite{devlin2017neural,bunel2018leveraging,chen2018execution} acts on a 2D grid world that contains an actor, various markers, and obstacles. The Karel DSL contains conditions and loops, which are not part of the DeepCoder DSL. However, the latter has some additional high-level functions, such as sort or map.~\citet{bunel2018leveraging} have demonstrated how to perform RL for improving the results on this DSL.~\citet{chen2018execution} have contributed the encoding of the current execution state (similar to~\citet{zohar2018automatic}), as well as the usage of ensembles. In our work, we do not study the effect of ensembles.

\begin{figure*}[t]
\vskip 0.2in
\begin{center}
\begin{tabular}{c@{\quad\quad\quad}c}
\includegraphics[width=.86945\columnwidth]{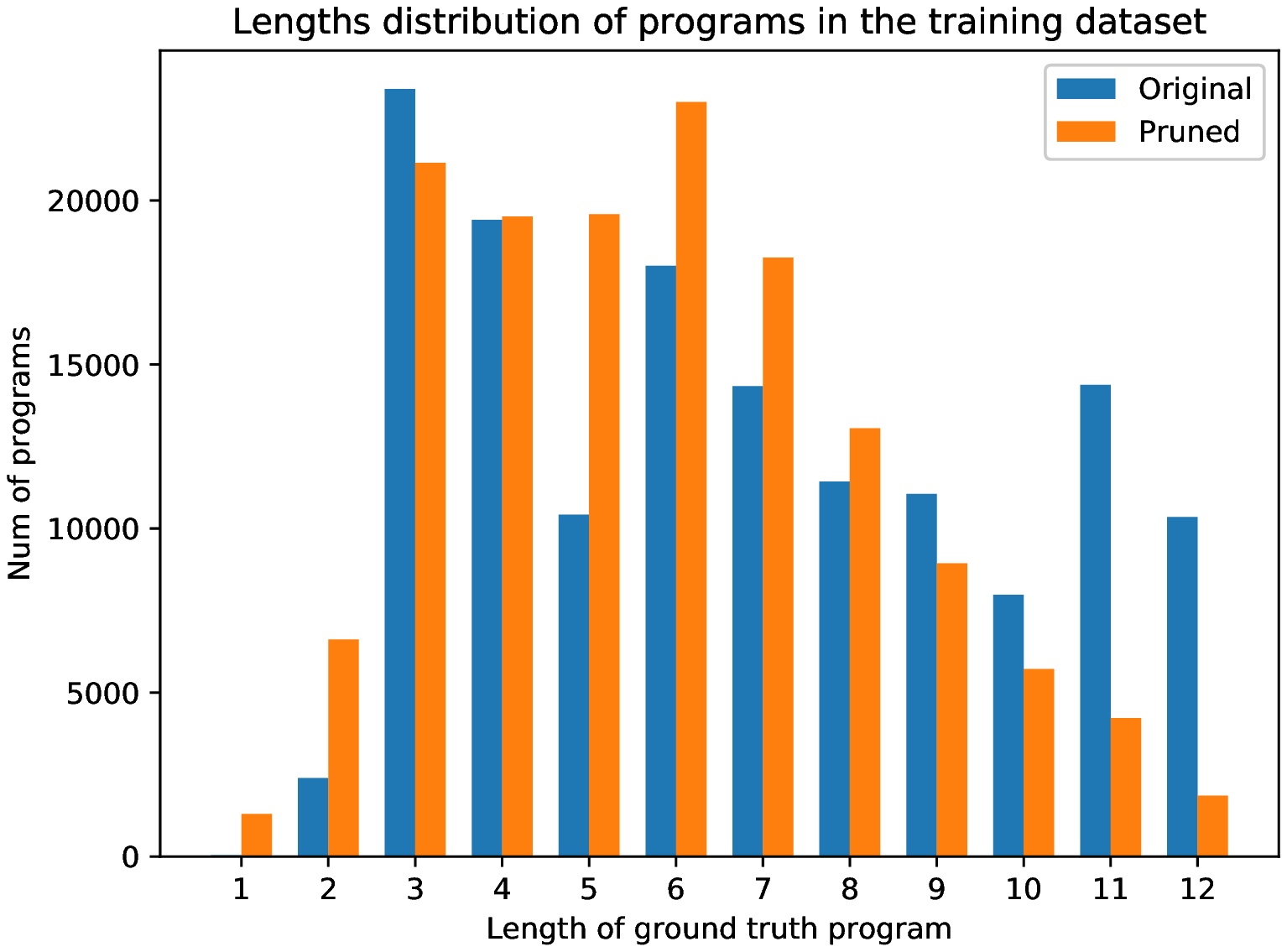}&
\includegraphics[width=.86945\columnwidth]{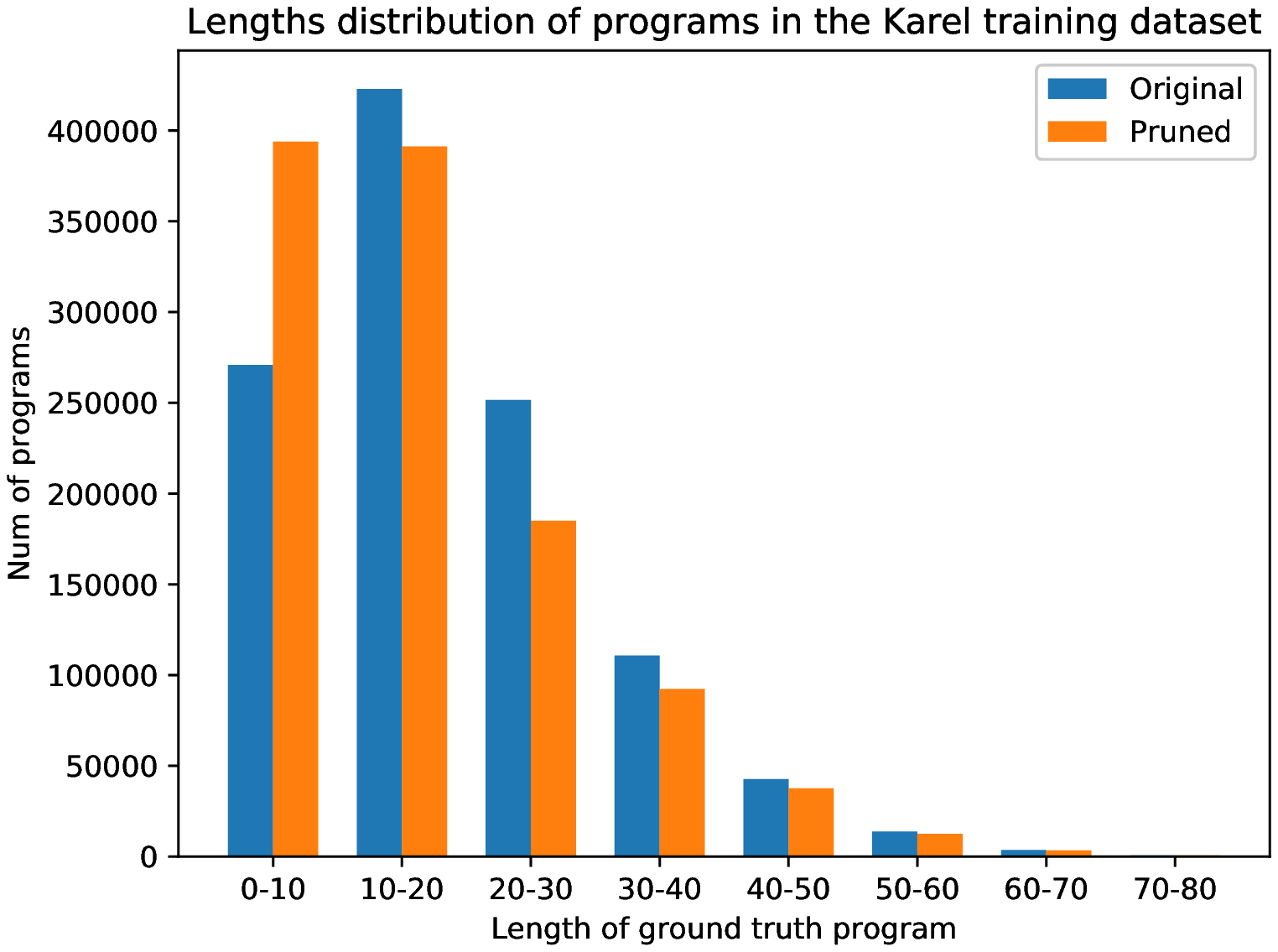}\\
%(a)&(b)\\
\end{tabular}
\vskip -0.2in
\caption{Length of ground truth programs in the training dataset before and after pruning. (Left) DeepCoder DSL. (Right) Karel.}
\label{fig:pruned-lengths-dist}
\end{center}
\vskip -0.2in
\end{figure*}

%\paragraph{Search methods} 
Beam search is perhaps the most popular method for exploring search trees. It is similar in nature to a best-first search, but instead of keeping in memory just the most promising candidate, beam search keeps several candidate tracks simultaneously. The number of candidates is the beam width, and the number of the derived expansions extracted from each candidate track  (prior to pruning back to the beam width) 
is the extraction width.

Complete Anytime Beam (CAB) search~\cite{cab} is an extension of beam search, which is better-suited for searching under a fixed time budget. In beam search, if the beam width and expansion size are configured too low, the beam search will be too weak to reach its destination. If they are too high, the beam search will reach the execution time limit, before achieving its goal. CAB extends beam search by running it repeatedly, starting with weak and cheap settings and increasing the search parameters after each search failure, as long as it still has time to run.

Monte Carlo Tree Search (MCTS)~\cite{chaslot2007progressive} is a popular method for prioritizing exploration in game search trees. The search is gradually expanded towards a mix of promising nodes and unexplored nodes. It includes the following steps: (1) Selection: traverse the search tree, until a new leaf node is reached. (2) Expansion: add the leaf node to the tree. (3) Simulation: evaluate the value of the leaf node (4) Backpropogation: update the values of the parent nodes on the track leading to the leaf node, according to its value. In our work, we do not employ a value function and the last two steps are not employed.

A major milestone in the field of artificial intelligence was the achievement of super-human performance in Go by using deep learning, reinforcement learning, and MCTS~\cite{silver2016mastering,silver2017mastering}.
Our method shares with it the specific fraction that is based on the visit count. However, we advocate for performing the count at the environment level, while in the previous Go work the count is based on the location in the tree and not on the board configuration. 

Our method does not use a learned or a simulation (rollout) based value function, as  done by~\citet{silver2017mastering}. We have tried learning such a value function and observed no improvement in the results. This may stem from the fact that the end goal, i.e., the output of our programs, varies, making such learning too challenging. In addition, as noted for the case of solving Rubik's cube with RL~\cite{mcaleer2018solving}, the value function mainly reduces the depth of the MCTS and not its breadth. In our case, since the length of the programs is limited, the depth is less of a concern.

\section{Method}

Our method is based on applying MCTS in lieu of beam search or CAB, which served as the search technique for previous work. We employ the same networks that are used in~\cite{zohar2018automatic}. An encoder $E$ embeds the input states of the execution, as well as the output for each sample and mean pooling is employed in order to obtain a single state vector. Based on this vector, a network $P$ predicts the next statement and a network $G$ predicts the variables to be dropped. $P$ and $G$ share most of their layers. In our work, we propose to add, as an additional input to the network $P$, the embedding of the partial program that was executed up to the current state, see Sec.~\ref{sec:pastembedding}.

\subsection{Our MCTS variant}
\label{sec:mcts}

Given a trained network $P$ that can prioritize the most likely next commands, we can explore the programs' space with a search tree. The nodes of the tree correspond to the execution states, after choosing the statements. The search favors nodes that have been less explored in the past (exploration), as well as nodes that result from commands that are predicted to have higher likelihood (exploitation). The score
%\begin{equation}
$U(s,a) = \frac{P(s,a)}{N(s,a)+1}$ is used,
%\end{equation}
where $P(s,a)$ is the output of the trained network (pseudo-probability of choosing action $a$, i.e., command, given state $s$), and $N(s,a)$ is the number of times we took action $a$, as a response to being in state $s$.

Unlike the literature we are aware of, in our variant of the search we do not backtrack. Instead, we start the search each time from the root node. At first,  we pick the most promising command according to the network's output, until we reach a solution, an invalid execution state, or a maximum program length limit. If a solution was not found, we start the search over, this time employing the updated score $U$, which discounts nodes that have been visited. 

\paragraph{Shared count} 

While the MCTS methods we are aware of count the number of visits per tree node, we found it useful to count the number of times each state is visited.

A state is the memory status obtained after running the partial program (from the root node to the current node) on each of the inputs in the input/output specifications. Two states are identical,  if the memory status is identical for all of the inputs in the specification.

In the DeepCoder DSL, the memory contains the various registers. We sort the registers such that two states are the same, even if the order of the registers is permuted, as long as the underlying values are the same. In the Karel DSL, the memory contains the status of the grid world.

\subsection{Pruning}

Since there are multiple execution pathways that result in the same states, learning with an MLE loss could be suboptimal. Previous work~\cite{bunel2018leveraging} has suggested employing RL to overcome this;  However, as noted in subsequent work, this does not necessarily lead to significantly improved performance. 

We suggest, instead, to preprocess the training set such that programs are replaced with shorter programs that are consistent with the input/output specifications. For this task, we employ a network that is pretrained on the original training data. If when applying this network, with our variant of MCTS, to the training dataset, a shorter (specification consistent) program is found, we replace the ground truth program. After this pruning, training is repeated from scratch.

Pruning has a drastic effect on the length of programs in the training datasets, as can be seen in Fig.~\ref{fig:pruned-lengths-dist}. For example, for the DeepCoder DSL, it shortens $82\%$ of the length 12 programs, which is the maximal length available in the training dataset. %In the Karel case, we also modify the definition of syntactic tokens to merge sequences of tokens that must appear together into a single supertoken.

\subsection{Encoding of Partial Programs}
\label{sec:pastembedding}

The input in PCCoder~\cite{zohar2018automatic} to the network $P$, which predicts the next statement, is the embedding of the current registers of the  execution environments (one environment per each sample input/output), including the end-goal (output) register of each sample. For Karel, we employ a similar architecture to PCCoder. In this case, $P$ receives an embedding of the input/output grids and an embedding of the last command executed, where the latter part was added following~\citet{bunel2018leveraging}.

In this work, we propose to add an additional input, which encodes the previous commands executed by the partial program that has already been constructed during the search.

For the DeepCoder DSL, the partial program is given as a sequence of 4-tuples containing an operator index, argument indices (one or two, if the operator takes one argument, we duplicate it), and an output index. We apply embedding layers on these four indices. 

LSTM is then applied to the concatenation of the four embeddings. The number of hidden units is taken to be the sum of the number of options in each look-up-tables we employ for the index embedding (66). The hidden state at the end of the sequence is then concatenated to the activations that arise from the other inputs of $P$, after the 5 densenet layers of $P$ are applied to these inputs. Two fully-connected layers (of the same size) are then applied, followed by a softmax layer of the same dimensionality, as in the original $P$.

In Karel, the input and output grids are encoded, using the same convolutional networks used by~\cite{bunel2018leveraging}. The partial program is encoded by running two LSTMs: the first is run on the embedding of the commands (there are no arguments) and the second on the nesting level (is the command executed within an if condition, a while loop, etc). The LUTs contain 38 command options and six nesting options. The LSTMs have a corresponding number of hidden units. The hidden states of the two LSTMs at the end of the sequence are concatenated to the embedded input/output samples (after the CNN encoding) and passed through three linear layers, with 512 hidden units. This is done separately for each input/output sample, and is aggregated by max pooling, followed by a softmax layer.

\section{Experiments}

We test the various search methods and evaluate the effect on performance of each of our contributions. In all of our experiments, the PCCoder architecture is used. Applying this method for Karel involved the same input/output encoding used in~\cite{bunel2018leveraging}. Since, unlike this previous work, we do not use an LSTM to generate the program (PCCoder predicts only the next command), the LSTM is replaced with a fully connected network. A garbage collection network $G$ is not required in this DSL.

Note that given enough time, one can find, using a na\"{\i}ve search approach,  the programs, up to a certain length, that are consistent with the input/output pairs. The shortest such program can then be selected, in order to improve generalization. Therefore, it is important to enforce a maximal allowed run time. This has obvious drawbacks when comparing across contributions. However, given the same standard hardware, a fixed runtime budget provides a reliable performance metric. In our experiments, we employ a commercial cloud infrastructure to perform all experiments.

The DeepCoder DSL experiments were done with eight registers and the benchmark of~\cite{zohar2018automatic} that includes training programs of lengths up to 12, and ground truth test programs of length 14 (all test environments are constructed with programs of length 14, which may or may not have shorter equivalents).

\begin{table}[t]
\caption{Accuracy on the DeepCoder benchmark (percents). The pruned column states whether the training set was pre-pruned. PartEnc states whether the encoding of the partial program was added to the prediction network $P$. Three search strategies are compared (beam search is not competitive), where shared means MCTS with shared counts. Different timeout values are tested. $^1$ These results correspond to the results of~\cite{zohar2018automatic}.}
\label{tab:deepcoderresults}
\centering
\begin{small}
\begin{tabular}{lllccc}
\toprule
    & & & \multicolumn{3}{c}{Timeout}\\
    \cmidrule(lr){4-6}
	Pruned & PartEnc & Search & 500 & 1000 & 2000 \\
	\midrule
	No & No & CAB$^1$ & 39.4 & 44.2 & 50.2 \\ 
	 &  & MCTS & 59.4 & 65.2 & 69.6 \\ 
	 &  & shared & 67.0 & 72.4 & 76.0 \\ 
	 \midrule
	No & Yes & CAB & 54.4 & 59.6 & 65.2 \\ 
	 &  & MCTS & 69.6 & 75.4 & 80.2 \\ 
	 &  & shared & 72.2 & 78.8 & 83.0 \\ 
	 \midrule
	Yes & No & CAB & 43.8 & 48.8 & 53.6 \\ 
	 &  & MCTS & 59.0 & 63.2 & 68.8 \\ 
	 &  & shared & 64.8 & 69.0 & 72.6 \\ 
	 \midrule
	Yes & Yes & CAB & 62.4 & 65.8 & 71.8 \\ 
	 &  & MCTS & 74.4 & 79.0 & 82.2 \\ 
	 &  & shared & 77.4 & 81.2 & 85.2 \\ 
	 \bottomrule
\end{tabular}
\end{small}
\end{table}

\begin{table}[t]
\caption{Accuracy on the Karel benchmark (precents) MCTS+ and Shared+ indicates the addition of both dataset pruning and partial program encoding to the search method.}
\label{tab:karelresults}
\centering
\begin{small}
\begin{tabular}{ccccc}
\toprule
	 & \multicolumn{2}{c}{Timeout=1000} & \multicolumn{2}{c}{Timeout=2000}  \\ 
	 \cmidrule(lr){2-3}
	 \cmidrule(lr){4-5}
	 Method & 5-goal & Incl. held-out & 5-goal & Incl. held-out \\ 
	 \midrule
	Beam & 86.28 & 81.00 & 86.28 & 81.00 \\ 
	CAB & 88.24 & 82.72 & 89.40 & 83.60 \\ 
	MCTS & 87.24 & 81.76 & 88.96 & 83.08 \\ 
	MCTS+ & 89.20 & 83.04 & 90.08 & 83.68 \\ 
	Shared & 87.16 & 80.96 & 87.84 & 81.48 \\ 
	Shared+ & 89.04 & 83.00 & 89.88 & 83.64 \\ 
	\bottomrule
\end{tabular}
\end{small}
\end{table}

The results are given in Tab.~\ref{tab:deepcoderresults}. As can be seen, all of our contributions (pruning of the dataset, encoding the partial program, replacing CAB with our variant of MCTS, and the shared count MCTS) improve performance over the baseline. This is true for the three different values of timeout. Our best result, which includes all contributions, presents an accuracy of 85.2\%, in comparison to 50.2\% of the previous work, for the same timeout of 2000 seconds. Partial program encoding by itself adds 15\% to the CAB search. Pruning has a less dramatic effect: it adds 3\% to CAB. 

In addition, pruning does not add to the shared visit count MCTS method by itself. However, when combined with encoding of the partial history, it adds significantly, e.g., 5.2\% for the 500 second timeout.

In all four groups obtained by applying or not applying pruning or partial program encoding, the MCTS method with the shared counts outperforms the other search method by a significant margin, and our variant of MCTS without the shared counts outperforms CAB.

The Karel benchmark used is taken from~\cite{devlin2017neural} and is also used by~\cite{bunel2018leveraging, chen2018execution}. However, 
on this benchmark, we found it challenging to compare directly with previous work, since the previous work did not employ a timeout. Instead, a fixed number of beams was used, which is not compatible with our MCTS variant going back to the root at each failed search. 

Performance is measured by exposing five input/output samples to the searcher and measuring if the searcher succeeded in finding a program which is not only consistent with the five input/output examples, but also generalizes to a sixth unseen input/output sample. In Tab.~\ref{tab:karelresults}, we present accuracy for both goals (consistency with the five or the six samples). The results are given for timeouts of 1000 or 2000 seconds. Beam search results are given for the best width and expansion parameters that we have found (128 and 5, respectively). The CAB parameters were also optimized to maximize the test accuracy, and starting with a similar beam search, it doubles the beam width and increases the expansion size by one, at each repeated attempt.

For this DSL, the combination of MCTS with past encoding and dataset pruning obtains the best results in both timeouts and for both success criteria. Sharing the visit counts between identical world grids did not improve results and the added bookkeeping increased runtime, leading to slightly lower results.

\section*{Acknowledgement}
This work was supported by an ICRC grant.

\bibliography{synth}
\bibliographystyle{icml2019}

\end{document}